\documentclass[sigconf]{acmart}
\AtBeginDocument{%
  }

\copyrightyear{2024} 
\acmYear{2024} 
\setcopyright{rightsretained} 
\acmConference[ICCAD '24]{IEEE/ACM International Conference on Computer-Aided Design}{October 27--31, 2024}{New York, NY, USA}
\acmBooktitle{IEEE/ACM International Conference on Computer-Aided Design (ICCAD '24), October 27--31, 2024, New York, NY, USA}
\acmDOI{10.1145/3676536.3676694}
\acmISBN{979-8-4007-1077-3/24/10}

\usepackage{anyfontsize}
\usepackage{soul}
\usepackage{url}
\usepackage[utf8]{inputenc}
\usepackage{graphicx}
\usepackage{amsmath}
\usepackage{amsthm}
\usepackage{booktabs}
\usepackage{algorithm}
\usepackage[switch]{lineno}
\usepackage{textcomp}
\usepackage{booktabs} 
\usepackage{pifont}
\usepackage{soul}
\usepackage{comment}
\usepackage{algorithmicx}
\usepackage{algpseudocode}
\usepackage{mathtools}
\usepackage{multirow}
\usepackage{colortbl}
\newcommand{\cmark}{{\color{green}\ding{51}}}
\newcommand{\xmark}{{\color{red}\ding{55}}}

\usepackage{stmaryrd}

\newcommand{\share}[1]{\llbracket #1 \rrbracket}
\newcommand{\bshare}[1]{\langle #1 \rangle}

\usepackage{tikz}

\newcommand*\circled[1]{\tikz[baseline=(char.base)]{
            \node[shape=circle,draw,inner sep=0.1pt] (char) {#1};}}
            
\usepackage{xspace}          
\newcommand{\ouralg}{AdaPI\xspace}

\begin{document}

\title{\texttt{\ouralg}: Facilitating DNN Model \underline{Ada}ptivity for Efficient \underline{P}rivate \underline{I}nference in Edge Computing}

\author{Tong Zhou$^{*1}$, Jiahui Zhao$^{*2}$, Yukui Luo$^3$, Xi Xie$^2$, Wujie Wen$^4$, Caiwen Ding$^2$ and Xiaolin Xu$^1$}

\affiliation{$*$Equal Contribution $^1$Northeastern University\country{} $^2$University of Connecticut, $^3$UMass Dartmouth, $^4$NC State University}
\email{{zhou.tong1, x.xu}@northeastern.edu, {jiahui.zhao, xi.xie, caiwen.ding}@uconn.edu, yluo2@umassd.edu, wwen2@ncsu.edu}

\begin{abstract}
Private inference (PI) has emerged as a promising solution to execute computations on encrypted data, safeguarding user privacy and model parameters in edge computing. However, existing PI methods are predominantly developed considering constant resource constraints, overlooking the varied and dynamic resource constraints in diverse edge devices, like energy budgets. Consequently, model providers have to design specialized models for different devices, where all of them have to be stored on the edge server, resulting in inefficient deployment. 
To fill this gap, this work presents \texttt{\ouralg}, a novel approach that achieves adaptive PI by allowing a model to perform well across edge devices with diverse energy budgets. \ouralg employs a PI-aware training strategy that optimizes the model weights alongside weight-level and feature-level soft masks. These soft masks are subsequently transformed into multiple binary masks to enable adjustments in communication and computation workloads. Through sequentially training the model with increasingly dense binary masks, \ouralg attains optimal accuracy for each energy budget, which outperforms the state-of-the-art PI methods by 7.3\% in terms of test accuracy on CIFAR-100. 
The code of \ouralg can be accessed via \url{https://github.com/jiahuiiiiii/AdaPI}.

\end{abstract}

\maketitle

\section{Introduction}
Bringing computation closer to data sources, edge computing reduces network congestion and achieves faster response, benefiting numerous real-time applications, including those based on deep neural network (DNN) inference \cite{shi2016edge, zhou2024application, zhou2024portfolio, lai2023detect, lai2024selective, shen2024harnessing, zhang2024cu, weng2024fortifying, weng2024big, wang2024ceb, jiang2024catching, mathavan2023inductive, yuan2024label, jiang2024media, yu2024credit, zheng2024advanced, cao2024rough,mo2024fine, liu2024enhanced, xu2024investigating,lin2024text, elhedhli2017airfreight,jin2024apeer, jin2024learning, he2024networkalignmenttransferablegraph, zhou2024exploring}. 
However, user privacy is a major concern in this context, as edge devices would continuously collect and process data containing sensitive information, e.g., users' location and personal data \cite{cao2020overview}. Additionally, safeguarding high-performance DNN models, which are valuable intellectual property owned by edge computing service providers, is crucial to prevent unauthorized usage and illegal extraction \cite{zhou2023nnsplitter,zhou2022obfunas,liu2023mirrornet}.

To ensure secure edge computing, private inference (PI), such as multi-party computation (MPC), has been envisioned as a promising solution \cite{knott2021crypten,ghodsi2020cryptonas,huang2022cheetah}. This approach enables DNN computation on ciphertext, involving encrypted input and model parameters.
For example, MPC employs cryptographic primitives (e.g., secret sharing \cite{beimel2011secret, peng2023pasnet, peng2023autorep, luo2023aq2pnn, duan2024ssnet}) to realize PI \cite{hua2020guardnn, peng2024lingcn}. Unfortunately, it introduces significant computation and communication overhead, resulting in extended response times and additional energy consumption \cite{mishra2020delphi}. 
Within a 2-party computation setting involving the edge server and the edge client (a configuration adopted in subsequent sections), conducting inference via ResNet-50 on cipher-text results in latency overhead surging up to 50 times in comparison to plaintext computation, with ReLU accounting for over 99\% of this latency increase \cite{peng2023pasnet}. 
Given that simply removing ReLU can lead to a drastic drop in inference accuracy, several methods have been proposed to reduce PI latency by replacing ReLU with linear operations \cite{cho2022selective, jha2021deepreduce} or polynomial approximation \cite{mishra2020delphi, lou2020safenet}.

\begin{figure}[t]
    \centering \includegraphics[scale=1.8]{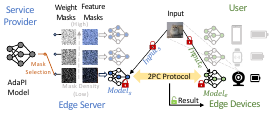}
    \caption{The inference process for \ouralg. For the edge device with low energy budget, the weight and feature masks with low density are chosen for the server-side model. The masked model is also encrypted and installed on the edge device. The inference process is divided between the edge server and the device, ensuring user privacy and model confidentiality.}
    \label{fig:illustration}
\end{figure}
 
While earlier efforts in private inference (PI) have shown promising ReLU-accuracy balance \cite{cho2022selective, jha2021deepreduce,mishra2020delphi, lou2020safenet}, they cannot inherently achieve efficient deployment in edge computing. This limitation arises from their primary emphasis on
optimizing model accuracy under a static communication or computation workload (associated with a fixed energy budget). 
However, as illustrated in Fig. \ref{fig:illustration}, PI entails collaborative computation involving both edge servers and devices \cite{peng2023pasnet}, with energy budgets varying across devices due to factors like battery capacities \cite{gusev2018going}.   
Consequently, two notable challenges arise when applying existing PI methods directly.
First, model providers must craft tailored DNNs for distinct devices to accommodate diverse energy budgets, which is time-consuming and requires substantial engineering efforts. 
Second, all these models have to be stored in the edge server to support PI, resulting in storage efficiency. 
Thus, it is critical to enable the adaptivity of DNN models to facilitate efficient PI in edge computing, which has been largely unexplored in previous PI works.

To fill this gap, we aim to optimize a DNN model to adapt to varying energy budgets across diverse edge devices. 
Recognizing that energy consumption is influenced by both communication and computation factors~\cite{garimella2023characterizing}, we aim to calibrate these workloads to align with the specific energy constraints of the deployed devices. 
Specifically, (i) we adjust the communication workload by utilizing feature masks associated with different ReLU densities, as ReLU operations prominently contribute to the communication workload within PI scenarios. 
Also, (ii) we adjust the numbers of Multiply-Accumulate (MAC) operations to allow different computation workloads, which can be determined by weight masks characterized by varying densities. 
Thus, the edge server can only store one model alongside multiple masks to efficiently interact with different edge devices to enable PI, as illustrated in Fig. \ref{fig:illustration}.

However, it is non-trivial to enable PI adaptivity in edge computing, due to the following challenges: (i) While achieving either a balance between ReLU and accuracy or between MAC and accuracy is relatively straightforward~\cite{cho2022selective,gong2022all}, striking a balance among accuracy, ReLU density, and MAC density is challenging, especially because prioritizing MAC efficiency may risk compromising ReLU efficiency~\cite{jha2021deepreduce}; 
(ii) Each mask can potentially interfere with others while sharing a single set of model weights; 
(iii) This interference complicates the optimization process, making it difficult to maximize the model accuracy for each mask (associated with a certain energy budget) without adversely impacting other masks.

We propose a novel PI-aware model training approach, namely \ouralg, to address these challenges. To tackle the first challenge, \ouralg formulates a triple optimization problem to achieve a balance among these factors.
Besides, \ouralg utilizes a weight-level soft mask and a feature-level soft mask along with an indicator function to address the second challenge. This indicator function converts optimized soft masks into binary masks based on desired densities associated with different energy levels. Furthermore, we propose a sequential training strategy for weight optimization to better preserve inference accuracy for each mask. Our proposed \ouralg addresses the limitations of existing methods and delves into the practical challenges faced in edge computing.

The contributions of this work are summarized as follows:
\vspace{-\topsep}
\begin{itemize}
    \item We introduce \ouralg, a novel approach facilitating model adaptivity for efficient PI based on a sequential multi-mask training strategy. The adaptivity minimizes the necessity for extensive reconfiguration efforts, ensuring secure edge inference across diverse energy budgets.
    
    \item We propose soft masks featuring indicator functions to address a triple optimization problem. This problem-solving mechanism strikes a balance between accuracy, computation workload, and communication workload.
    
    \item We propose a unified metric to facilitate a more comprehensive comparison of model performance under triple optimization with prior PI methods.
    
    \item Through extensive experiments, we demonstrate the effectiveness of \ouralg in achieving adaptivity through mask selection, and the test accuracy can surpass SOTA PI methods around 10 times. We open source the code via a \href{https://anonymous.4open.science/r/AdaPI-4724/README.md}{\textit{hyperlink}}.
\end{itemize}

\section{Related Works}
\subsection{DNN in Edge Computing}

Edge computing has gained significant attention for its ability to provide services with low latency. It offers advantages over cloud computing by reducing the need for high-bandwidth connections and lowering data transfer costs through the offloading of services from the cloud to edge networks \cite{shi2016edge}. This makes it particularly suitable for time-sensitive applications relying on DNN models, such as smart conveyance and connected health, where faster response times and real-time decision-making are crucial \cite{covi2021adaptive}.

However, edge servers typically have limited computational resources and storage capacity compared to cloud servers. Such constraints require DNN models used for inference in edge computing to be lightweight, which can be achieved by model pruning techniques \cite{liu2020pruning}. 
Moreover, the hardware resources vary for different edge devices, including power, storage capacity, and communication capabilities. To allow a model suitable for various edge devices, the model should be adaptive to different resource constraints \cite{islam2022eve,gong2022all}. 

In addition to resource constraints, edge computing also presents challenges related to model confidentiality and data privacy. By bringing computation closer to the data source, the risk of model leakage and data leakage also increases. To ensure a secure ML service, PI techniques can be leveraged for DNN models, addressing the need for confidentiality and protecting data privacy. 

\subsection{Secret-Sharing-Based MPC}
In this work, we explore a two-party secure computing (2PC) protocol \cite{kamara2011outsourcing} to enable PI in edge computing. 
By partitioning inference to different entities, MPC performs computations without revealing individual inputs (i.e., user data and model weights).  As illustrated in Fig. \ref{fig:illustration}, the input and model are securely shared between the edge user and the edge server. The edge device computes $input_e$ and $model_e$, and the server computes $input_s$ and $model_s$, where $model_e$ and $model_s$ share the same masks but with different weight values.
Here we introduce critical operations and cryptographic primitives of the 2PC protocol involved in PI for DNNs.

\noindent\textbf{Secret Sharing.}
Secret sharing is the most critical operation in MPC, which bridges the communication between parties while keeping one's information (users' data and model weights) secure without the risk of being extracted by other parties. Specifically, in this work, we adopt the commonly used secret sharing scheme described in CrypTen~\cite{knott2021crypten}. 
We denote the two secret shares as $\share{x}\gets(x_{p_1}, x_{p_2})$, where $x_{p_i}$ represents the share distributed to party $i$.
Here we outline the the share generation and the share recovering utilized in our approach: 

\begin{itemize}
    \item {\it Share Generation} $\mathbb{\textrm{shr}} (x)$: A random value $r$ in $\mathbb{Z}_{m}$ is sampled, and shares are generated as $\share{x}\gets (r, x-r)$.
    \item {\it Share Recovering}  $\mathbb{\textrm{rec}} ({\share{x}})$: Given $\share{x}\gets (x_{p_1}, x_{p_2})$, it computes $x\gets x_{p_1} + x_{p_2}$ to recover $x$.
\end{itemize}

\begin{table*}[htp]
    \centering
    \resizebox{0.98\linewidth}{!}{
    \begin{tabular}{l|c|c|c|c|l}
    \hline
    \hline
     & \textbf{Privacy} & \textbf{Comp. Redu.}$^\star$ &  \textbf{Commu. Redu.}$^{\diamond}$ & \textbf{Adaptivity} * & \textbf{Techniques}  \\ \hline
     DELPHI \cite{mishra2020delphi} & \cmark & \xmark &  \cmark & \xmark & Hybrid cryptographic protocol + NAS\\
     CryptoNAS \cite{ghodsi2020cryptonas} & \cmark & \xmark & \cmark & \xmark  & NAS + ReLU pruning/shuffling\\
        SNL \cite{cho2022selective} & \cmark & \xmark& \cmark& \xmark & ReLU linearization via $l_1$-penalty\\
        SafeNet \cite{lou2020safenet}& \cmark&  \xmark& \cmark& \xmark & Multi-degree approximation in channel-wise\\
        SENet \cite{kundu2022learning} & \cmark & \xmark & \cmark & \xmark  & Three-stage training method\\
        EVE \cite{islam2022eve}& \xmark & \cmark & \xmark & \ding{109} & Pattern pruning + NAS \\
        Once-For-All \cite{caionce} & \xmark & \cmark & \xmark& \ding{109}   & NAS + progressive shrinking algorithm  \\
        All-in-One \cite{gong2022all} & \xmark & \cmark & \xmark&  \ding{109}   & Parametric pruning + switchable BatchNorm\\
        
        \textbf{\ouralg ~(this work)} & \cmark& \cmark&\cmark & \cmark & Triple optimization + soft masks + sequential training\\
        \hline
        \multicolumn{6}{l}{Note: $\star$ Denotes computation reduction  \quad \quad   * Include computation adaptive and communication adaptive  }\\
\multicolumn{6}{l}{\quad \quad  $\diamond$ Denotes communication reduction  \quad \; \ding{109} Denotes only being computation adaptive.}\\
        \hline
    \end{tabular}}
\caption{The comparison of previous works with our proposed method.}
\label{tab:comparison}
\end{table*}

\textbf{Secure Multiplication.}
We denote secret shared matrices as $\share{X}$ and $\share{Y}$ and consider the use of matrix multiplicative operations in the secret-sharing pattern, i.e., $\share{R}\gets \share{X} \otimes \share{Y}$. Here $\otimes$ represents a general multiplication such as Hadamard product, matrix multiplication, and convolution. To generate the required Beaver triples~\cite{beaver1991efficient} $\share{Z}=\share{A}\otimes\share{B}$, we employ an oblivious transfer (OT)~\cite{kilian1988founding} based approach, with $A$ and $B$ initialized randomly. 
Subsequently, each party computes two intermediate matrices, $E_{p_i} = X_{p_i} - A_{p_i}$ and $F_{p_i} = Y_{p_i} - B_{p_i}$, separately. The intermediate shares are then jointly recovered, with $E\gets \mathbb{\textrm{rec}} {(\share{E})}$ and $F\gets \mathbb{\textrm{rec}} {(\share{F})}$. Finally, each party $p_i$ locally calculates the secret-shared $R_{p_i}$ to get the result:
\begin{equation}\label{eq:mat_mul_ss}
R_{p_i} = -i\cdot E \otimes F + X_{p_i} \otimes F  + E \otimes Y_{p_i} + Z_{p_i}
\vspace{-3pt}
\end{equation}
\textbf{Secure 2PC Comparison.}
This protocol, also known as the millionaires' protocol, is designed to determine which of two parties holds a larger value, without revealing the actual value to each other. We use the same protocol as CrypTen~\cite{knott2021crypten} to conduct comparison ($\share{X < 0}$) through (1) arithmetic share $\share{X}$ to binary share $\bshare{X}$ conversion, (2) right shift to extract the sign bit $\bshare{b} = \bshare{X} >> (L-1)$ ($L$ is the bit width), and (3) binary share $\bshare{b}$ to arithmetic share $\share{b}$ conversion for final evaluation result.

Overall, these cryptographic primitives provide secure computation, which can safeguard user privacy and model parameters in edge computing. However, compared to plain-text computing, the complex computation of encrypted data introduces a substantial computational workload. Besides, evaluating non-linear operations such as ReLU—which activates only if the input is positive (i.e., greater than zero)—necessitates secure comparison. For this, values initially represented in arithmetic shares must be converted into binary shares. This conversion typically involves multiple rounds of interaction between the parties to securely compute the bits of the original value, which incurs a significant communication workload.

\subsection{Limitation of Prior Works}
The requirements for secure and efficient DNN inference in edge computing encompass privacy and adaptability to diverse energy budgets, through reductions in communication and computation. Yet, prior studies have either delved into adaptivity for on-device inference by adjusting computation workload or solely concentrated on optimizing communication workload for a fixed budget, falling short of meeting all requirements, as summarized in Tab. \ref{tab:comparison}.

Several efforts have been made to make a DNN model adaptive to diverse devices \cite{caionce} or devices with dynamic resource constraints by employing weight pruning to adjust computation workload \cite{islam2022eve,gong2022all}. 
For instance, the once-for-all network is proposed to support diverse architectural settings \cite{caionce}, where it first trains a full net and then progressively fine-tunes to support smaller sub-networks. Importantly, they have to fine-tune both large and small sub-networks to avoid interfering, incurring substantial training costs.  In contrast, our sequential multi-mask training allows the model to maximize performance for every mask without requiring any fine-tuning.
More importantly, these methods did not optimize ReLU efficiency, which is the primary bottleneck in PI tasks. Consequently, they would be inefficient when applying secure computing protocols to achieve secure edge computing.

To ensure data privacy and model confidentiality, prior works have utilized PI techniques and optimized the communication workload by reducing ReLU counts \cite{cho2022selective,jha2021deepreduce,mishra2020delphi,lou2020safenet}.
Given that simply removing ReLU can lead to a drastic drop in inference accuracy, 
\cite{cho2022selective}  selectively replace ReLU with linear operation using a gradient-based algorithm while maintaining inference accuracy. 
Moreover, \cite{peng2023autorep}  introduces distribution-aware polynomial approximation to accurately approximate ReLUs, achieving a better trade-off between ReLU density and inference accuracy. 
However, these methods yield dense model weights, contributing to a burdensome computation workload. Additionally, they solely optimize a model for a fixed ReLU/communication budget, limiting their efficient deployment across diverse devices.

\section{Proposed Approach: \ouralg}
\begin{figure*}
    \centering
    \includegraphics[width=0.9\linewidth]{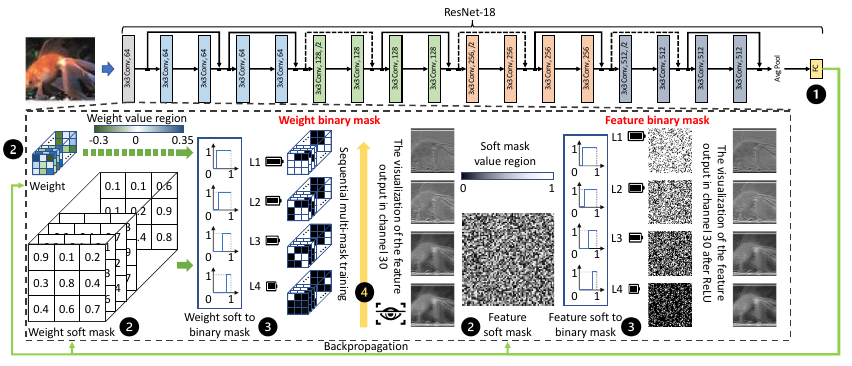}
    \caption{Overview of \ouralg with ResNet-18. The process involves the following steps:  
    (1) Calculation of the loss using Eq. \ref{eq:problem}. (2) Optimization of weights and soft masks with the lowest weights/ReLU budgets through a triple optimization problem using the backpropagation method. (3) Conversion of soft masks to binary masks (where L1-L4 represent different levels of energy budgets, the white region in masks indicating weights/ReLU operations are preserved). (4) Sequential training with masks from low density (associated with L4) to high density (associated with L1)}
    \label{fig:overview}
\end{figure*}

The proposed \ouralg achieves adaptive PI for secure and efficient DNN inference in
edge computing by solving the following problems: (i) How to mutually optimize inference accuracy, computation workload, and communication workload?  (ii) How to optimize multiple masks (associated with diverse computation/communication workloads) for a model without interfering with each other?  (iii) How to optimize the model weights so that the accuracy can be maximized for each mask? The solutions are detailed in Sec.~\ref{sec:soft}, Sec.~\ref{sec:optimization}, and Sec.~\ref{sec:training}, respectively, with overview shown in Fig. \ref{fig:overview}.

\subsection{Notation and Definition}

Let us consider a DNN model (denoted as $f$) with $L$ layers, parameterized by $\mathbf{W}$ = \{$W^0$, $W^1$, ... $W^L$\}. 
We use $\sigma(\cdot)$ to represent the ReLU operation and $Z^l \in \mathbf{Z}$ to denote the output of the $l$-th layer (referred to as the pre-activation feature maps). The feed-forward propagation can be described using the following equation:{\small
\begin{equation}
X^l = \sigma(Z^l) = \sigma(W^{l-1} * X^{l-1}),
\label{eq:notation}
\end{equation}
}
where 
$X^l$ represents the inputs of the $l$-th layer. For simplicity, we omit the bias term in each layer. The element-wise ReLU operation $\sigma(\cdot)$ acts on the feature level, specifically on each element $z_{i,j}$ in $Z^l$, where $(i,j)$ denotes the index of a 2-dimensional feature map. It is defined as:
{\small
\begin{equation}
\label{eq:relu}
  \sigma(z_{i,j}) = 
  \begin{cases}
  z_{i,j}, & \textnormal{if} ~ z_{i,j} > 0 \\
  0, & \textnormal{otherwise}.
  \end{cases}
\end{equation}
}Conventional DNN model training involves weight optimizations  following Eq. (\ref{eq:conv_loss}):
{\small
\begin{equation}
\label{eq:conv_loss}
\min_\mathbf{W} \mathcal{L}(f(X^0_{tr}, \mathbf{W}), Y_{tr}),
\end{equation}
}
where $(X^0_{tr}, Y_{tr})$ represents the labeled training (tr) dataset, and $\mathcal{L}$ denotes the cross-entropy loss function used for image classification tasks.
However, using the DNN model trained with Eq. (\ref{eq:conv_loss}) directly is inefficient for privacy-preserving inference in edge computing. This is attributed to its dense MACs and non-linear ReLU operations, necessitating the optimization of both computation and communication workloads while maintaining accuracy.

\subsection{Triple Optimization}
\label{sec:optimization}

To align with energy budgets in computation, we adjust the DNN model's MAC operations by optimizing model sparsity through weight pruning.
A common strategy for weight pruning is to introduce an $l_0$ norm regularization term on the weights, which can simultaneously achieve weight pruning and preserve accuracy. This strategy can be implemented by adding an additional term to the loss function in Eq. (\ref{eq:conv_loss}).  However, directly pruning on weights will cause an imbalance of the loss term and the $l_0$ term as the latter will decrease in training, thus causing training instability \cite{xiao2019autoprune}. 
Instead of directly pruning the weights, we utilize a weight-level binary mask $M^b_w$  to decouple weight pruning from the training process: 
{\small
\begin{equation}
\label{eq:prune}
\min_{\mathbf{W},M^b_w} \mathcal{L}(f(X^0_{tr}, \mathbf{W} \odot M^b_w), Y_{tr}) + \lambda \mathcal{R}(M^b_w)
\end{equation}
}
By regularizing the sparsity of the mask via $\mathcal{R}(\cdot)$, we can achieve weight sparsity without causing training instability, so as to 
reduce computation workloads.

To reduce communication overhead, we adopt the linearization technique \cite{cho2022selective} for pruning ReLU operations. This involves replacing the ReLU activation function $\sigma(z_{i,j})$ defined in Eq. (\ref{eq:relu}) with $z_{i,j}$. Although this linearization reduces communication overhead, it also leads to accuracy degradation. To address this, we introduce a feature-level binary mask $M^b_r$ of the same size as $Z$, indicating which ReLU activation should be pruned to better preserve accuracy. The modified forward pass equation becomes:
{\small
\begin{equation}
X^l = \sigma(Z^l) \odot M^{b,l}_r + Z^l \odot (\textbf{1}-M^{b,l}_r)
\label{eq:forward}
\end{equation}
}

In order to achieve the triple objectives of computation reduction, communication reduction, and accuracy preservation, we train the model by optimizing the following equation:
{\small
\begin{equation}
\label{eq:single}
\min_{\mathbf{W},M^b_w,M^b_r} \mathcal{L}\left(f(X^0,\mathbf{W}\odot M^b_w,  M^{b}_r), Y\right) +\\
\lambda \mathcal{R}(M^b_w) + \mu \mathcal{R}(M^{b}_r),  
\end{equation}
}where $\mathcal{R}(\cdot)$ is the regularization term to control the sparsity of $M^b_w$ and $M^b_r$, which are associated with the weight density and ReLU density, respectively. Besides, the hyperparameters $\lambda$ and $\mu$ control the trade-off between computation/ communication reduction and accuracy maximization.
After optimization, the masks $M^b_w$ and $M^b_r$ can be applied to the model to meet a certain energy budget.


\subsection{Soft Mask with Indicator Function}
\label{sec:soft}
To achieve adaptivity, allowing a single set of weights to cater to multiple masks associated with various energy budgets, we aim to optimize multiple weight-level/feature-level masks. The primary challenge lies in the potential negative impact of different masks on each other, given that they are applied to a single model and influence its optimization. 

In addressing this challenge, we 
first unify these masks during optimization, then convert them into multiple binary masks during inference. Specifically, we enhance \ouralg by introducing soft masks in place of binary masks, i.e., a weight-level soft mask $M^s_w$ and a feature-level soft mask $M^s_r$. These masks are trainable variables with floating-point values, providing flexibility in adjusting the level of computation workload and communication workload.
  
We also introduce the
indicator function $h$, which operates element-wise on the soft mask and converts it into a binary mask based on the threshold $\theta$. The indicator function is defined as:
{\small
\begin{equation}
h(m_{i,j}) = 
\begin{cases}
  1, & \textnormal{if} ~ m_{i,j} > \theta \\
  0, & \textnormal{otherwise}
  \end{cases}
\label{eq:indicator}
\end{equation}
}
where $\theta$ is the threshold associated with the weight/ReLU density.  For example, if the desired weight density is 0.5, then $\theta$ will be set to the value corresponding to the  50\% percentile of the weight range of each layer. 
In practical scenarios, the desired weight/ReLU density can be determined by profiling the edge device's communication \& computation capability and corresponding execution energy consumption. 

With the introduction of the soft mask and the indicator function, the problem formulation becomes:
{\small
\begin{equation}
\label{eq:problem}
\begin{split}
\min_{\mathbf{W},M^s_w,M^s_r} \mathcal{L} & \left(f(X^0,\mathbf{W}\odot h(M^s_w),  h(M^s_r), Y\right) +\\
 &\lambda \mathcal{R}(h(M^s_w)) + \mu  \mathcal{R}( h(M^s_r))
\end{split}
\end{equation}
}Once we obtain the optimized soft masks $M^s_w$ and $M^s_r$, we can convert them into multiple binary masks by adjusting the threshold $\theta$. This adjustment is based on the computation and communication budgets, which can be determined by profiling the devices' practical communication and computation capabilities in practical. A larger threshold leads to a sparser binary mask, while a smaller threshold results in a denser binary mask.
These binary masks are nested, meaning that the weights and ReLU operations preserved in a sparser mask are also preserved in denser masks. This is intuitive since the critical weights and ReLU operations that have the most significant impact on accuracy are preserved across all masks.

\subsection{Sequential Multi-Mask Training}
\label{sec:training}
We propose a sequential multi-mask training strategy to maximize the accuracy for each binary mask. The strategy involves initially optimizing the model with soft masks linked to the lowest density, which ensures that the model performs well with the sparsest mask after converting soft masks to binary masks (line~\ref{line:init}-\ref{line:seek} in Alg.~\ref{alg:ouralg}). Subsequently, we gradually train additional weights associated with denser masks, as described in line~\ref{line:loop_s}-\ref{line:loop_end} in Alg.~\ref{alg:ouralg}. 
The intuition behind this is that the most sparse mask, which corresponds to the most stringent energy budget, would result in the lowest inference accuracy since more weights and ReLU are pruned. If we can achieve satisfactory accuracy with this mask, we can leverage the additional parameters and more ReLU operations associated with denser masks to further improve accuracy.

However, a challenge arises as the indicator $h(\cdot)$ in Eq. (\ref{eq:problem}) produces binary outputs (0 or 1), which are non-differentiable and cannot be directly trained using back-propagation. To overcome this challenge, we utilize the softplus-based straight-through estimator (STE) to estimate the gradient of the indicator \cite{xiao2019autoprune}. 
By replacing the non-differentiable indicator function with the softplus ($f(x) = log(1+e^x)$) during the backward pass, we can effectively train the entire network using backpropagation. 
In addition, we leverage knowledge distillation to further enhance the model performance. In knowledge distillation, the un-pruned model serves as the teacher model, which is trained using Eq. (\ref{eq:conv_loss}). To facilitate knowledge transfer, we employ the KL-divergence loss \cite{hinton2015distilling} and a peer-wise normalized feature map difference penalty \cite{zagoruykopaying}, which improves the performance of the model with different masks.

The overall training process is outlined in Alg.~\ref{alg:ouralg}. Through the combination of sequential multi-mask training and knowledge distillation, we achieve the training of an adaptive and high-performing model capable of accommodating various energy budgets.

\begin{algorithm}[tb!]
\caption{Sequential Multi-mask Training in \ouralg}
\label{alg:ouralg}
\begin{algorithmic}[1]
\Require Training data $(X, Y)$, weights densities $d_w$, ReLU densities $d_r$ 
\State Initialize model weights $\mathbf{W}$
\State Train teacher model $f_t$ \label{line:teacher} \Comment{Eq. (\ref{eq:conv_loss})} 

\State Initialize soft masks $M^s_w$ and $M^s_r$ \label{line:init}
\State Set $\theta_w$ and $\theta_r$ based on $min(d_w)$ and $min(d_r)$ \Comment{Eq. (\ref{eq:indicator})} 
\State Optimize $\mathbf{W}$ with $M^s_w$ and $M^s_r$ till converge \Comment{Eq. (\ref{eq:problem})} \label{line:seek}
\For{ $d_w$, $d_r$ in ascending order} \label{line:loop_s}
\State Determine $\theta_w$  based on $d_w$
\State Generate a weight binary mask via $h(M^s_w)$ with $\theta_w$ 
\State Determine $\theta_r$  based on $d_r$ 
\State Generate a feature binary mask via $h(M^s_r)$ with $\theta_r$ 
\State Optimize W with these two binary masks
\EndFor \label{line:loop_end}
\State \Return Model weights $\mathbf{W}$ with multiple binary masks
\end{algorithmic}
\end{algorithm}

\section{Unified Metric}

Previous works primarily focus on optimizing ReLU and employ ReLU count as their evaluation metric~\cite{cho2022selective,jha2021deepreduce}. Different, \ouralg addresses both MAC and ReLU reductions, necessitating a unified metric for comparison. Therefore, we opt to convert MACs into ReLU counts and propose \textit{Normalized ReLU count} as the unified metric for our experimental evaluation. This metric includes both the MAC-converted ReLU counts and the original ReLU counts.
Here, we present our latency modeling approach and demonstrate how we normalize the MACs into ReLU counts using latency modeling. 

\subsection{Latency of 2PC-Conv Operator} 
The 2-party Convolution (2PC-Conv) operator involves multiplication in ciphertext. The computation part follows tiled architecture implementation~\cite{zhang2015optimizing}. 
There are four tiling parameters ($Tm$, $Tn$, $Tc$, $Tr$) that correspond to the input channel, output channel, column, and row tile. Tiling parameters can be adjusted according to memory bandwidth and on-chip resources to reduce the communication-to-computation (CTC) ratio and achieve better performance. 

Assuming we can meet the computation roof by adjusting tiling parameters, the latency of the 2PC-Conv computation part (considering density as $D$) can be estimated as 
\begin{equation}
\small
CMP_{\text{Conv}} = \frac{3 \times K \times K \times FO^2 \times IC \times OC}{PP \times freq} \times D
\end{equation}
where $K$ is the convolution kernel size, $IC$ and $OC$ denote the number of input channels and output channels, and the output feature is square with size $FO$. We denote the computational parallelism as $PP$. The communication latency is modeled as $ COMM_{Conv} = T_{bc} + \frac{32 \times FI^2 \times IC}{Rt_{bw}}$, where $T_{bc}$ denotes two-party network build connection time and $Rt_{bw}$ is the effective network bandwidth between server and client. 
Thus, the latency of 2PC-Conv is: 
\begin{equation}\label{eq:Ltconv}
    Lat_{2PC-Conv} = CMP_{Conv} + 2 \times COMM_{Conv}
\end{equation}

\subsection{Latency of 2PC-ReLU Operator}
Since 2PC-ReLU requires \textbf{2PC-OT Processing Flow}, we provide the communication detail of OT-based comparison protocol~\cite{kilian1988founding}. 
Assume both parties have a shared prime number $m$, one generator ($g$) selected from the finite space $\mathbb{Z}_m$, and an \textbf{index} list with $L$ length. As we adopt 2-bit part, the length of \textbf{index} list is $L=4$.

\noindent\textbf{\circled{1}} \textbf{Server} ($S_0$) generates a random integer $rd_{s_0}$, and compute mask number $S$ with $S = g^{rd_{S_0}}\ mod\ m$, then shares $S$ with the Client ($S_1$). 
We only need to consider communication ($COMM_1$) latency as $COMM_1 = T_{bc} + \frac{32}{Rt_{bw}}$, since computation ($CMP_1$) latency is trivial.

\noindent\textbf{\circled{2}} \textbf{Client} ($S_1$) received $S$, and generates ${R}$ list based on $S_1$'s 32-bit dataset ${M_1}$, and then send them to $S_0$. Each element of ${M_1}$ is split into $U = 16$ parts, thus each part is with 2 bits. 
Assuming the input feature is square with size $FI$ and $IC$ denotes the input channel, and we denote the computational parallelism as $PP$. The $CMP_2$ is modeled as Eq.~(\ref{eq:CMP_2}), and $COMM_2$ is modeled as Eq.~(\ref{eq:COMM_2}). 
\begin{equation}\label{eq:CMP_2}
\small
    CMP_2 = \frac{32 \times 17 \times FI^2 \times IC}{PP \times freq}
\end{equation}   
\begin{equation}\label{eq:COMM_2}
\small
    COMM_2 = T_{bc} + \frac{32 \times 16 \times FI^2 \times IC}{Rt_{bw}}
\end{equation}

\noindent\textbf{\circled{3}} \textbf{Server} ($S_0$) received ${R}$ and will first generate the encryption ${key_0}(y,u) = {R}(y,u)\oplus (S^{b2d({M_1}(y, u)) + 1}\ mod\ m)^{rd_{S_0}}\ mod\ m$. The $S_0$ also generates its comparison matrix for its ${M_0}$ with a 32-bit datatype and $U = 16$ parts. Thus, the matrix size for each value ($x$) is $4 \times 16$. The encrypted $Enc({M_0}(x,u))={M_0}(x,u)\oplus {key_0}(y,u)$ will be sent to $S_1$. The $COMM_3$ of this step is shown in Eq.~(\ref{eq:COMM_3}), and $CMP_3$ can be estimated as Eq.~(\ref{eq:CMP_3}). 
\begin{equation}\label{eq:CMP_3}
\small
    CMP_3 = \frac{32 \times (17 + (4 \times 16)) \times FI^2 \times IC}{PP \times freq}
\end{equation}    
\begin{equation}\label{eq:COMM_3}
\small
    COMM_3 = T_{bc} + \frac{32 \times 4 \times 16 \times FI^2 \times IC}{Rt_{bw}}
\end{equation}

\noindent\textbf{\circled{4}} \textbf{Client} ($S_1$) decodes the encrypted massage by ${key_1} = S^{rd_{S_0}}\ mod\ m$ in the final step.  
The $CMP_4$ and $COMM_4$ are calculated as following:
\begin{equation}\label{eq:CMP_4}
\small
    CMP_4 = \frac{((32 \times 4 \times 16) + 1) \times FI^2 \times IC}{PP \times freq}
\end{equation}    
\begin{equation}\label{eq:COMM_4}
\small
    COMM_4 = T_{bc} + \frac{FI^2 \times IC}{Rt_{bw}}
\end{equation}
Therefore, the 2PC-ReLU latency ($Lat_{2PC-ReLu}$) model is defined:
\begin{equation}\label{eq:LtReLU}
\small
    Lat_{2PC-ReLU} = \sum_{i = 2}^{4}CMP_i + \sum_{j = 1}^{4}COMM_j
\end{equation}

\textbf{ReLU Normalization.}
Using latency modeling proposed in Eq.~(\ref{eq:Ltconv}) and Eq.~(\ref{eq:LtReLU}), we can effectively normalize the MACs count into ReLU count by matching the MACs-related latency in 2PC-Conv operator with ReLU induced latency in 2PC-ReLU operator. The ReLU normalization could bridge the gap between weight compression and ReLU reduction, thus introducing a systematic view of accelerating MPC-based private inference on the edge platforms.

\section{Experiments}
In our experiments, we estimate the energy consumption associated with the model under different workloads, showing its capability to accommodate diverse devices with varying energy budgets. Besides, we compare \ouralg with SOTA methods to demonstrate that our solution can achieve better performance in terms of test accuracy.

\subsection{Experimental Setup}
\label{sec:setup}
\textbf{Architectures and Datasets:} following the SOTA work SNL \cite{cho2022selective}, we evaluate \ouralg using ResNet-18 \cite{he2016deep} and WideResNet-22-8 \cite{zagoruyko2016wide} on three datasets, including CIFAR-10/CIFAR-100 \cite{krizhevsky2009learning} and Tiny-ImageNet \cite{chrabaszcz2017downsampled} datasets. 
We limit our selection to these architectures and datasets as more complex DNNs or datasets are better suited for cloud computing rather than edge computing applications.

\noindent\textbf{Hardware Setup:}
We utilize a two-party MPC setup for PI, employing two ZCU104 Multi-Processor System-on-Chip (MPSoC) platforms as a case study for evaluation. One platform serves as the edge server, while the other acts as the user. Both platforms are connected to a router through a Local Area Network (LAN) with a bandwidth capacity of $Rt_{bw} = 1 GB/s$.
Considering a 128-bit load/store bus width and 32-bit data, we concurrently load and store four data units while implementing the kernel at a frequency of freq = 200 MHz. 
For the two-party computation (2PC) inference, we adopt a similar protocol described in CrypTen \cite{knott2021crypten}, and set the fixed-point ring size as 64 bits \cite{knott2021crypten}.

\noindent\textbf{Adaptivity Setup:} 
Given our contribution lies at the algorithmic level, we conduct evaluations on a single hardware platform with different energy budgets, representing diverse devices, to efficiently validate the feasibility and effectiveness of our method. 
We simulate four levels of computation and communication workloads, each associated with specific weight/ReLU densities compared to the full model. 
These levels include L1 (0.4), L2 (0.2), L3 (0.1), and L4 (0.05). While we experiment with consistent weight/ReLU densities for simplicity, it is crucial to note that \ouralg supports arbitrary densities for weights and ReLU.
When applying \ouralg for diverse devices in the real world, we can profile these devices to get their communication \& computation capability and corresponding execution energy consumption, then configure model ReLU and weight density level to satisfy the given energy budgets.

\noindent
\textbf{Hyper-parameters Settings:}
When implementing line \ref{line:seek} in Alg. \ref{alg:ouralg} with the weights/ReLU density of L4, the AdamW optimizer is utilized with learning rate $LR = 0.001$, $\beta_1 = 0.9$, and $\beta_2 = 0.999$ for model weight and both soft masks, as well as an additional $10^{-4}$ weight decay for the model weight. The training encompasses 250 epochs, resulting in the final soft masks and a sparsified model. 
Beginning with L4, we incrementally raise the density level during sequential training. At each level, the sparsified model is trained using stochastic gradient descent with a learning rate of $LR = 0.01$ and a cosine annealing LR scheduler. Training is conducted for 300 epochs on the CIFAR-10 and CIFAR-100 datasets and 160 epochs for the Tiny-ImageNet dataset at each density level.
\begin{figure*}[tb!]
    \centering
      \includegraphics[width=.9\linewidth]{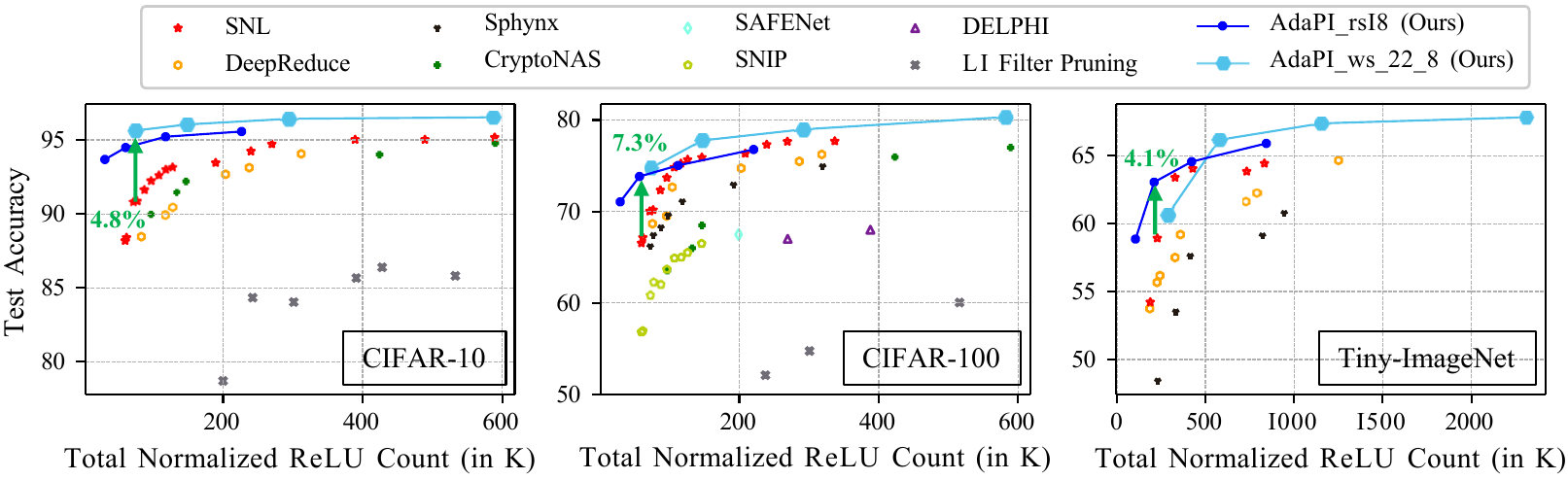}
  \captionof{figure}{For each dataset, \ouralg achieves Pareto frontiers of the normalized ReLU count vs. test accuracy with a single set of weights, with the density level from left to right being L4, L3, L2, and L1. In contrast, SOTA methods design specialized models for every scenario. AdaPI\_rs18 and AdaPI\_ws\_22\_8 denote \ouralg on ResNet-18 and WideResNet-22-8, respectively.
  }
    \label{fig:Pareto_frontiers}
\end{figure*}

\noindent
\textbf{Teacher Models:} \ouralg includes a teacher model for each architecture and dataset, aiming to better preserve the inference accuracy of the adaptive model by leveraging knowledge distillation (line \ref{line:teacher} in Alg. \ref{alg:ouralg}). The performance of teacher models is provided in Tab.~\ref{tab:teacher}.

\begin{table}[tb!]

\centering
\resizebox{\linewidth}{!}{
\begin{tabular}{cccccc}
\toprule
\textbf{Dataset} & \textbf{Model} & \textbf{Weights (M)} & \textbf{MACs (M)} &\textbf{ ReLU (K)} & \textbf{Accuracy (\%)}\\ \midrule
\multirow{2}{*}{CIFAR-10} & ResNet & 11.16 & 555.42 & 491.52  & 96.04\\
& WideResNet & 17.15 & 2454.11 & 1359.87  & 96.73\\ \midrule
\multirow{2}{*}{CIFAR-100} & ResNet & 11.21 & 555.42 &  491.52  & 78.42\\
& WideResNet & 17.19 & 2454.11 & 1359.87 & 81.02 \\ \midrule
\multirow{2}{*}{Tiny-ImageNet} & ResNet & 11.26 & 2221.77  & 1966.08  &  66.94 \\
& WideResNet & 17.24  & 9816.54 & 5439.49  &  68.38\\ \bottomrule

\end{tabular}
}
\caption{The statistics and performance of teacher models.}
\label{tab:teacher}
\end{table}

\subsection{Comparison Methods}
\ouralg is the first that introduces adaptive PI to diverse energy budgets from both communication and computation aspects, there are no existing works that can be directly used as comparison methods. Instead, we compare our approach with SOTA PI techniques and weight pruning methods to demonstrate its effectiveness and performance. 
The PI works we include for comparison are SNL \cite{cho2022selective}, DeepReDuce \cite{jha2021deepreduce}, Sphynx \cite{cho2022sphynx}, CryptoNAS \cite{ghodsi2020cryptonas}, SAFENet \cite{lou2020safenet}, and DELPHI \cite{mishra2020delphi}. For weight pruning, we include SNIP \cite{lee2018snip} and L1 filter pruning \cite{li2016pruning} in our comparison. It is important to note that these works do not enable adaptivity, so they have to generate multiple models for different weights/ReLU densities. In contrast, \ouralg only generates one model for multiple masks to achieve adaptivity.

We provide a detailed comparison with SNL since we experiment with the same DNNs (i.e., ResNet-18 and WideResNet-22-8). \textit{We did not include SNL's energy consumption as it uses a different hardware setting (CPU + GPU frameworks) and their energy data is not reported.}

\begin{table}[tb!]
\centering
\resizebox{0.98\linewidth}{!}{

\begin{tabular}{cccccc}
\toprule
\multicolumn{1}{c}{\textbf{Methods}} & \textbf{\begin{tabular}[c]{@{}c@{}}MACs (M)\end{tabular}} & \textbf{\begin{tabular}[c]{@{}c@{}}Normalized \\ReLU (K)\end{tabular}} & \textbf{\begin{tabular}[c]{@{}c@{}}Accuracy\\ (\%)\end{tabular}} & \textbf{\begin{tabular}[c]{@{}c@{}}Latency\\ (s)\end{tabular}}
& \textbf{\begin{tabular}[c]{@{}c@{}}Energy\\ (J)\end{tabular}}\\ \midrule
\multicolumn{5}{c}{ResNet-18} \\ \midrule
\multicolumn{1}{c}{SNL} & 555.4 & 95.96 & 73.75 & 1.07 & N/A \\
\multicolumn{1}{c}{SNL} & 555.4 & 59.49 & 66.53 & 0.29 & N/A \\
\multicolumn{1}{c}{\textbf{AdaPI (L1)}} & 285.3 & 220.54 & 76.78 & 0.56 & 5.1\\
\multicolumn{1}{c}{\textbf{AdaPI (L2)}} & 167.0 & 112.32 & 75.03 & 0.33 & 3.0\\
\multicolumn{1}{c}{\textbf{AdaPI (L3)}} & 92.1 & 56.83 & 73.82 & 0.22 & 2.0\\
\multicolumn{1}{c}{\textbf{AdaPI (L4)}} & 52.9 & 29.01 & 71.05 &  0.16 & 1.4\\ \midrule
\multicolumn{5}{c}{WideResNet-22-8} \\ \midrule
\multicolumn{1}{c}{SNL} & 2454.1 & 269.01 & 77.65 & 4.05 & N/A\\
\multicolumn{1}{c}{SNL} & 2454.1 & 208.47 & 76.35 & 2.80 & N/A\\
\multicolumn{1}{c}{\textbf{AdaPI (L1)}} & 1042.0 & 581.9 & 80.33 & 1.41 & 12.7\\
\multicolumn{1}{c}{\textbf{AdaPI (L2)}} & 549.7 & 292.0 & 78.98 &  0.80 & 7.2\\
\multicolumn{1}{c}{\textbf{AdaPI (L3)}} & 287.2 & 146.4 & 77.76 &  0.50 & 4.5\\
\multicolumn{1}{c}{\textbf{AdaPI (L4)}} & 152.5 & 73.54 & 74.75 & 0.35 & 3.1\\ \bottomrule
\end{tabular}}
\caption{The performance of \ouralg on CIFAR-100 with comparison of SNL.} 
\label{tab:cifar100}
\end{table}

\subsection{Results on CIFAR-10/100}
We compare \ouralg with the SOTA methods on CIFAR-10/100, as illustrated in Fig. \ref{fig:Pareto_frontiers}.
Regarding CIFAR-10, \ouralg achieves Pareto frontiers of the normalized ReLU count and test accuracy for both ResNet-18 and WideResNet-22-8 models. Specifically, when experimenting with ResNet-18 on CIFAR-10, \ouralg achieves an impressive accuracy of 94.49\% with a normalized ReLU count of 60K. Remarkably, this corresponds to a ReLU density and weight density of just 10\% compared to the full model, i.e., the teacher model. In contrast, under a similar normalized ReLU count, the SNL method only achieves about 88\% accuracy. Besides, \ouralg with WideResNet-22-8 can achieve a 4.8\% improvement in test accuracy compared to SNL under a similar normalized ReLU count.

As for the results on CIFAR-100, it is evident that \ouralg with ResNet-18 attains significantly higher test accuracy with a smaller normalized ReLU count compared to other SOTA methods. Specifically, \ouralg achieves up to 7.3\% higher accuracy compared to SNL, with details reported in Tab. \ref{tab:cifar100}. 
Moreover, when using WideResNet-22-8, \ouralg surpasses all SOTA methods. Notably, with fewer MACs and ReLUs,  the latency can be reduced by about 5 times from L1 to L4 (1.41s vs. 0.35s). 
Based on the estimated energy consumption in Tab.~\ref{tab:cifar100}, by selecting different masks associated with varying weights/ReLU densities, \ouralg can accommodate devices with varying energy budgets. 
When facing more stringent resource constraints, the test accuracy may slightly decrease, reaching around 74.74\%. However, if more computation and communication resources are available, the accuracy can achieve 80.33\%.

Besides, to better understand how communication workload can be reduced via ReLU pruning, we provide detailed statistics in Tab.~\ref{tab:comm}. For example, by reducing the ReLU density from 0.40 to 0.05, the communication
volume of using ResNet-18 on CIFAR-100 is decreased by 3.6$\times$ ( from 58.83MB to 16.26MB). 

\begin{table}[tb!]
\centering
\resizebox{0.98\linewidth}{!}{
\begin{tabular}{ccccc}
\toprule
\multirow{3}{*}{\textbf{Methods}} & \multicolumn{2}{c} {\textbf{CIFAR-100}} &
\multicolumn{2}{c} {\textbf{Tiny-ImageNet}}\\
    \cmidrule(r){2-3} \cmidrule(r){4-5}
    & \textbf{\begin{tabular}[c]{@{}c@{}}ReLU (K)\\ /Density\end{tabular}} & \textbf{\begin{tabular}[c]{@{}c@{}}Communication\\ Volume (MB)\end{tabular}} & \textbf{\begin{tabular}[c]{@{}c@{}}ReLU (K)\\ /Density\end{tabular}} & \textbf{\begin{tabular}[c]{@{}c@{}}Communication\\ Volume (MB)\end{tabular}}\\ \midrule
\multicolumn{5}{c}{ResNet-18} \\ \midrule
\multicolumn{1}{c}{\textbf{AdaPI (L1)}} & 196.61/0.40 & 58.83 & 786.43/0.40 & 235.31 \\
\multicolumn{1}{c}{\textbf{AdaPI (L2)}} & 98.30/0.20 & 34.51 & 393.22/0.20 & 138.00 \\
\multicolumn{1}{c}{\textbf{AdaPI (L3)}} &  49.15/0.10 & 22.33 & 196.61/0.10 & 89.35 \\
\multicolumn{1}{c}{\textbf{AdaPI (L4)}} & 24.58/0.05 & 16.26 & 98.30/0.05 & 65.02 \\ \midrule
\multicolumn{5}{c}{WideResNet-22-8} \\ \midrule
\multicolumn{1}{c}{\textbf{AdaPI (L1)}} & 543.95/0.40 & 154.18 & 2175.80/0.40  & 616.66 \\
\multicolumn{1}{c}{\textbf{AdaPI (L2)}} & 271.97/0.20 & 86.86 & 1087.90/0.20&  347.43\\
\multicolumn{1}{c}{\textbf{AdaPI (L3)}} & 135.99/0.10 & 53.21 & 543.95/0.10 &  212.80\\
\multicolumn{1}{c}{\textbf{AdaPI (L4)}} & 67.99/0.05 & 36.38 & 271.97/0.05 & 145.48 \\ \bottomrule
\end{tabular}}
\caption{Communication volume for diverse ReLU densities.}
\label{tab:comm}
\end{table}

\subsection{Results on Tiny-ImageNet}

To evaluate \ouralg on a larger dataset, we conducted experiments on Tiny-ImageNet, which contains 50k more training samples compared to CIFAR-10/100. In the figure on the right side of Fig. \ref{fig:Pareto_frontiers}, we can observe that, unlike the experiments on CIFAR-10/100 where WideResNet-22-8 consistently outperformed ResNet-18, WideResNet-22-8 under the stringent energy budget (L4) exhibits lower test accuracy on Tiny-ImageNet. This indicates that WideResNet-22-8 is parameter-efficient on Tiny-ImageNet and is more sensitive to weight pruning and ReLU removal, resulting in a larger drop in accuracy. 
However, when gradually transitioning from L4 to L1, WideResNet-22-8 demonstrates better performance than ResNet-18. 

We provide a detailed comparison with the SNL method in Tab. \ref{tab:tiny}. The outcomes underscore that under a comparable normalized ReLU count, \ouralg attains an accuracy exceeding that of SNL by more than 4.1\% higher accuracy than SNL (63.06\% vs. 58.94\%).
Overall, the experiments on Tiny-ImageNet demonstrate that \ouralg performs favorably compared to SNL and other SOTA methods. 

\begin{table}[tb!]
\centering
\resizebox{0.98\linewidth}{!}{
\begin{tabular}{cccccc}
\toprule
\textbf{Methods} & \textbf{\begin{tabular}[c]{@{}c@{}}MACs (M) \end{tabular}} & \textbf{\begin{tabular}[c]{@{}c@{}}Normalized \\ReLU (K)\end{tabular}} & \textbf{\begin{tabular}[c]{@{}c@{}}Accuracy\\ (\%)\end{tabular}} & \textbf{\begin{tabular}[c]{@{}c@{}}Latency\\ (s)\end{tabular}} & \textbf{\begin{tabular}[c]{@{}c@{}}Energy\\ (J)\end{tabular}}\\ \midrule
\multicolumn{5}{c}{ResNet-18} \\\midrule
SNL & 2221.8 & 328.88 & 61.65 & 7.77 & N/A\\
SNL & 2221.8 & 228.72 & 58.94 & 2.12 & N/A\\
\textbf{AdaPI(L1)} & 986.2 & 843.61 & 65.91 & 1.95 & 17.8\\
\textbf{AdaPI(L2)} & 508.6 & 422.70 & 64.57 & 1.07 & 9.6\\
\textbf{AdaPI(L3)} & 239.2 & 210.47 & 63.06 & 0.63 & 5.7\\
\textbf{AdaPI(L4)} & 123.6 & 105.47 & 58.87 & 0.42 & 3.7 \\ \midrule
\multicolumn{5}{c}{WideResNet-22-8} \\ \midrule
SNL & 9816.5 & 833.1 & 64.42 & 10.28 & N/A\\
\textbf{AdaPI(L1)} & 3881.4 & 2307.5 & 67.70 & 3.47 & 31.2\\
\textbf{AdaPI(L2)} & 1938.3 & 1153.7 & 67.38 & 1.96 & 31.2\\
\textbf{AdaPI(L3)} & 957.4 & 576.4 & 66.17 & 1.21 & 17.6 \\
\textbf{AdaPI(L4)} & 477.7 & 288.1 & 60.66 & 0.84 & 7.6 \\ \bottomrule
\end{tabular}}
\caption{The performance of \ouralg on Tiny-ImageNet with comparison of SNL.}
\label{tab:tiny}
\end{table}
\section{Discussion}

\begin{figure}[tb!]
    \centering    \includegraphics[width=.8\linewidth]{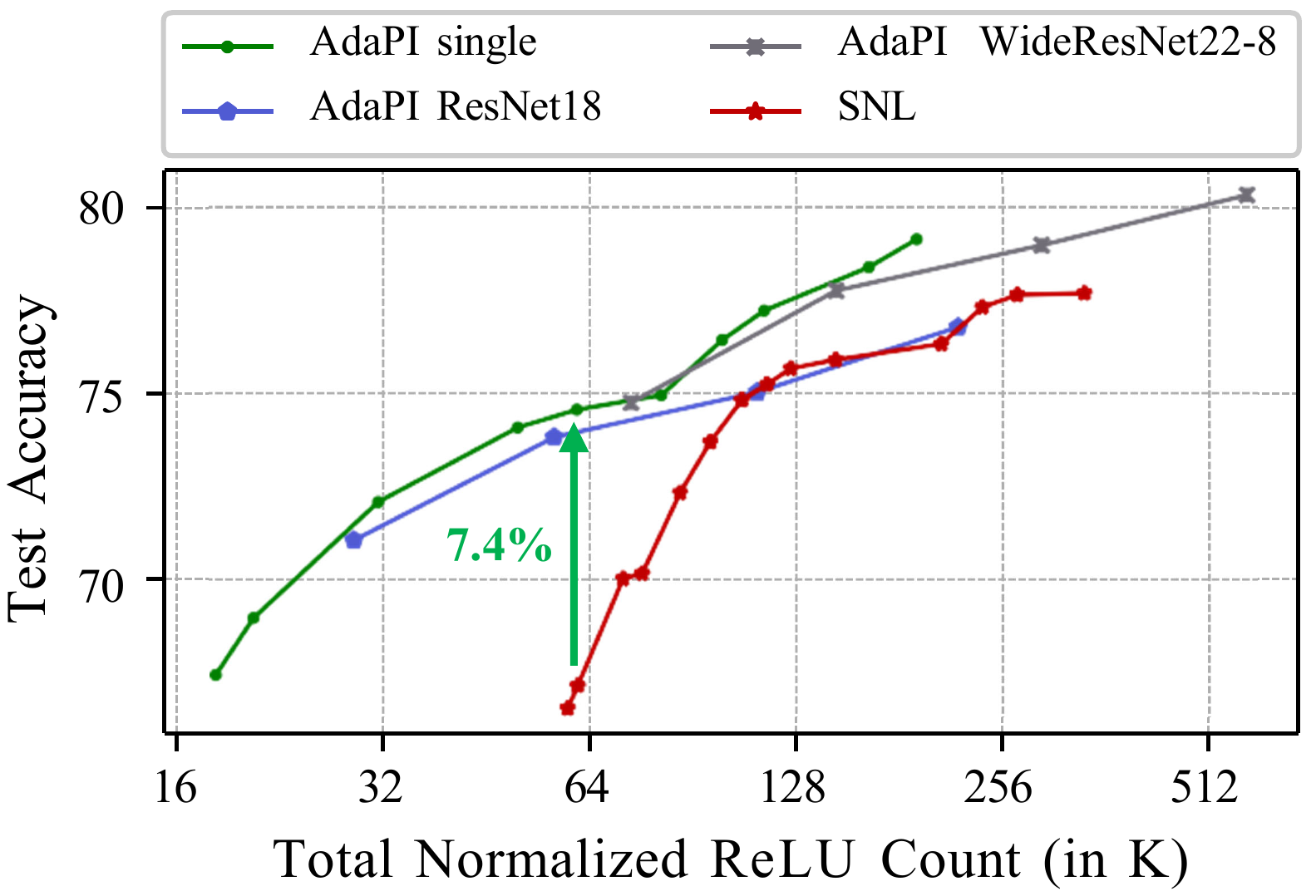}
    \caption{Performance comparison of \ouralg-single with \ouralg and SNL on CIFAR-100. 
    }
    \label{fig:wo_adp}
\end{figure}

We investigate the trade-off between accuracy and storage in the pursuit of adaptivity. 
To achieve adaptivity, we can optimize the model for each pair of masks ($M^b_r$ and $M^b_w$), following Eq. (\ref{eq:single}) (i.e., \ouralg-single) that involves generating multiple sets of model weights and masks to accommodate varying resource constraints. We examine its performance under various weight densities and ReLU densities on CIFAR-100 as an ablation study. The comparison of this method with \ouralg is shown in Fig. \ref{fig:wo_adp}, which indicates that \ouralg-single achieves higher accuracy compared to \ouralg with ResNet-18. This performance discrepancy arises from the nature of \ouralg-single, which tailors the model optimization for each mask associated with a distinct weight/ReLU density, where all optimized models have to be stored on the edge.  
On the other hand, employing WideResNet-22-8 within  \ouralg framework leads to comparable test accuracy across various ReLU counts with only a single set of weights. Overall, \ouralg strikes a balance between accuracy and deployment efficiency, thus facilitating PI in edge computing.

\section{Conclusion}
This paper presents \ouralg, a new method facilitating adaptive PI in edge computing. \ouralg accommodates a single set of model weights to varying energy budgets across diverse edge devices. Our method optimizes both feature-level and weight-level soft masks during model optimization. These masks transform into binary counterparts, adjusting communication and computation workloads. Thus, the edge server stores one model with multiple masks.
Through extensive experiments, we demonstrate that AdaPI has consistently outperformed SOTA methods in terms of test accuracy and the normalized ReLU count. In particular, on CIFAR-100 with ResNet-18, AdaPI achieves 7.3\% accuracy improvement compared to SNL. 
More importantly, by selecting suitable masks, \ouralg can conduct PI while incurring varying energy consumption, efficiently accommodating edge devices with diverse energy budgets.

\section{Acknowledgement}
This work is supported in part by the U.S. National Science Foundation under grants OAC-2319962, CNS-2239672, 2153690, 2326597, 2247892, 2348733, and 2247893.

\bibliographystyle{ACM-Reference-Format}
\bibliography{ref.bib}

\end{document}


\maketitle

\section{Secret-Sharing-Based MPC}
In this paper, we explore a two-party secure computing (2PC) protocol \cite{kamara2011outsourcing} to enable private inference (PI) in edge computing. This protocol lets the edge server and the edge device owned by users compute a function securely without revealing intermediate information or results.

\subsection{Secret Sharing}
Secret sharing is the most critical operation in multi-party computation, which bridges the communication between parties while keeping one's information (users' data and model weights) secure without the risk of being extracted by other parties. Specifically, in this work, we adopt the commonly used secret sharing scheme described in CrypTen~\cite{knott2021crypten}. 
As a symbolic representation, $\share{x}\gets(x_{p_1}, x_{p_2})$ 
denotes the two secret shares, where $x_{p_i}, i\in \{0,1\}$, is the share distributed to the party $i$. The share generation and the share recovering adopted in our work are shown below: 
\begin{itemize}
    \item {\it Share Generation} $\mathbb{\textrm{shr}} (x)$: A random value $r$ in $\mathbb{Z}_{m}$ is sampled, and shares are generated as $\share{x}\gets (r, x-r)$.
    \item {\it Share Recovering}  $\mathbb{\textrm{rec}} ({\share{x}})$: Given $\share{x}\gets (x_{p_1}, x_{p_2})$, it computes $x\gets x_{p_1} + x_{p_2}$ to recover $x$.
\end{itemize}
As most operators used in DNNs can be implemented through scaling, addition, multiplication, and comparison, here we provide an overview of these basic operations.

\subsection{Cryptographic Primitives}

\textbf{Scaling and Addition.} We denote secret shared matrices as $\share{X}$ and $\share{Y}$. The encrypted evaluation is given in Eq.~\ref{eq:mat_ad_ss}, where $a$ is the scaling factor. 
\begin{equation}\label{eq:mat_ad_ss}
\share{aX+Y}\gets(aX_{p_1}+Y_{p_1}, aX_{p_2}+Y_{p_2})
\end{equation}
\textbf{Multiplication.}
We consider the use of matrix multiplicative operations in the secret-sharing pattern, specifically $\share{R}\gets \share{X} \otimes \share{Y}$, where $\otimes$ is a general multiplication such as Hadamard product, matrix multiplication, and convolution. To generate the required Beaver triples~\cite{beaver1991efficient} $\share{Z}=\share{A}\otimes\share{B}$, we utilize an oblivious transfer (OT)~\cite{kilian1988founding} based approach, with $A$ and $B$ being randomly initialized. It is important to ensure that the shapes of $\share{Z}, \share{A}$, and $\share{B}$ match those of $\share{R}, \share{X}$, and $\share{Y}$, respectively, in order to align the matrix computation. Next, each party computes two intermediate matrices, $E_{p_i} = X_{p_i} - A_{p_i}$ and $F_{p_i} = Y_{p_i} - B_{p_i}$, separately. The intermediate shares are then jointly recovered, with $E\gets \mathbb{\textrm{rec}} {(\share{E})}$ and $F\gets \mathbb{\textrm{rec}} {(\share{F})}$. Finally, each party $p_i$ (the edge server and the edge user' device) calculates the secret-shared $R_{p_i}$ locally to get the result:
\begin{equation}\label{eq:mat_mul_ss}
R_{p_i} = -i\cdot E \otimes F + X_{p_i} \otimes F  + E \otimes Y_{p_i} + Z_{p_i}
\vspace{-3pt}
\end{equation}
\textbf{Secure 2PC Comparison.}
In the context of secure MPC, the 2PC comparison protocol, also known as the millionaires' protocol, is designed to determine which of two parties holds a larger value, without revealing the actual value to each other. We uses the same protocol as CrypTen~\cite{knott2021crypten} to conduct comparison ($\share{X < 0}$) through (1) arithmetic share $\share{X}$ to binary share $\bshare{X}$ conversion, (2) right shift to extract the sign bit $\bshare{b} = \bshare{X} >> (L-1)$ ($L$ is the bit width), and (3) binary share $\bshare{b}$ to arithmetic share $\share{b}$ conversion for final evaluation result.

Overall, these cryptographic primitives provide secure computation, which can safeguard user privacy and model parameters in edge computing. However, compared to plain-text computing, the complex computation on encrypted data introduce substantial computational workload. Besides, using secure 2PC comparison to complete ReLU operations incurring significant communication workload.

\section{Normalization}
\subsection{Normalized ReLU Count}


Here we present the latency modeling in the experiment setup, and how we normalize the MACs into ReLU counts using latency modeling. 

\textbf{2PC-Conv Operator} The 2-party Convolution (2PC-Conv) operator involves multiplication between ciphertext. The computation part follows tiled architecture implementation~\cite{zhang2015optimizing}. 
There are four tiling parameters ($Tm$, $Tn$, $Tc$, $Tr$) that correspond to the input channel, output channel, column, and row tile. Tiling parameters can be adjusted according to memory bandwidth and on-chip resources to reduce the communication-to-computation (CTC) ratio and achieve better performance. 
Assuming we can meet the computation roof by adjusting tiling parameters, the latency of the 2PC-Conv computation part (considering density as $D$) can be estimated as $CMP_{Conv} = \frac{3 \times K \times K \times FO^2 \times IC \times OC}{PP \times freq} \times D$, where $K$ is the convolution kernel size, $IC$ denotes the number of input channels, $OC$ denotes the number of output channels, and output feature is square with size $FO$. We denote the computational parallelism as $PP$. The communication latency is modeled as $ COMM_{Conv} = T_{bc} + \frac{32 \times FI^2 \times IC}{Rt_{bw}}$. Thus, the latency of 2PC-Conv is given in Eq.~\ref{eq:Ltconv}. 

\begin{equation}\label{eq:Ltconv}
    Lat_{2PC-Conv} = CMP_{Conv} + 2 \times COMM_{Conv}
\end{equation}

\textbf{2PC-OT Processing Flow} While OT-based comparison protocol has been discussed in~\cite{kilian1988founding}, we hereby provide other communication detail. 
Assume both parties have a shared prime number $m$, one generator ($g$) selected from the finite space $\mathbb{Z}_m$, and an \textbf{index} list with $L$ length. As we adopt 2-bit part, the length of \textbf{index} list is $L=4$. 

\noindent\textbf{\circled{1}} \textbf{Server} ($S_0$) generates a random integer $rd_{s_0}$, and compute mask number $S$ with $S = g^{rd_{S_0}}\ mod\ m$, then shares $S$ with the Client ($S_1$). 
We only need to consider communication ($COMM_1$) latency as $COMM_1 = T_{bc} + \frac{32}{Rt_{bw}}$, since computation ($CMP_1$) latency is trivial.

\noindent\textbf{\circled{2}} \textbf{Client} ($S_1$) received $S$, and generates ${R}$ list based on $S_1$'s 32-bit dataset ${M_1}$, and then send them to $S_0$. Each element of ${M_1}$ is split into $U = 16$ parts, thus each part is with 2 bits. 
Assuming the input feature is square with size $FI$ and $IC$ denotes the input channel, and we denote the computational parallelism as $PP$. The $CMP_2$ is modeled as Eq.~\ref{eq:CMP_2}, and $COMM_2$ is modeled as Eq.~\ref{eq:COMM_2}. 

\begin{equation}\label{eq:CMP_2}
    CMP_2 = \frac{32 \times 17 \times FI^2 \times IC}{PP \times freq}
    \vspace{-3pt}
\end{equation}   
\begin{equation}\label{eq:COMM_2}
    COMM_2 = T_{bc} + \frac{32 \times 16 \times FI^2 \times IC}{Rt_{bw}}
    \vspace{-3pt}
\end{equation}

\noindent\textbf{\circled{3}} \textbf{Server} ($S_0$) received ${R}$, it will first generate the encryption ${key_0}(y,u) = {R}(y,u)\oplus (S^{b2d({M_1}(y, u)) + 1}\ mod\ m)^{rd_{S_0}}\ mod\ m$. The $S_0$ also generates is comparison matrix for it's ${M_0}$ with 32-bit datatype and $U = 16$ parts, thus the matrix size for each value ($x$) is $4 \times 16$. The encrypted $Enc({M_0}(x,u))={M_0}(x,u)\oplus {key_0}(y,u)$ will be sent to $S_1$. The $COMM_3$ of this step is shown in Eq.~\ref{eq:COMM_3}, and $CMP_3$ can be estimated as Eq.~\ref{eq:CMP_3}. 

\begin{equation}\label{eq:CMP_3}
    CMP_3 = \frac{32 \times (17 + (4 \times 16)) \times FI^2 \times IC}{PP \times freq}
    \vspace{-3pt}
\end{equation}    
\begin{equation}\label{eq:COMM_3}
    COMM_3 = T_{bc} + \frac{32 \times 4 \times 16 \times FI^2 \times IC}{Rt_{bw}}
\end{equation}

\noindent\textbf{\circled{4}} \textbf{Client} ($S_1$) decodes the interested encrypted massage by ${key_1} = S^{rd_{S_0}}\ mod\ m$ in the final step. 
The $CMP_4$ and $COMM_4$ are calculated as following:

\begin{equation}\label{eq:CMP_4}
    CMP_4 = \frac{((32 \times 4 \times 16) + 1) \times FI^2 \times IC}{PP \times freq}
\end{equation}    
\begin{equation}\label{eq:COMM_4}
    COMM_4 = T_{bc} + \frac{FI^2 \times IC}{Rt_{bw}}
\end{equation}

\textbf{2PC-ReLU Operator} 2PC-ReLU requires 2PC-OT flow. 
2PC-ReLU latency ($Lat_{2PC-ReLu}$) model is given in Eq.~\ref{eq:LtReLU}.

\begin{equation}\label{eq:LtReLU}
    Lat_{2PC-ReLu} = \sum_{i = 2}^{4}CMP_i + \sum_{j = 1}^{4}COMM_j
\end{equation}

\subsection{ReLU Normalization}

Using latency modeling proposed in Eq.~\ref{eq:Ltconv} and Eq.~\ref{eq:LtReLU}, we can effectively normalize the MACs count into ReLU count by matching the MACs related latency in 2PC-Conv operator with ReLU induced latency in 2PC-ReLU operator. The ReLU normalization could bridge the gap between weight compression and ReLU reduction, thus introducing a systematic view of accelerating MPC-based private inference on the edge platforms. 


\section{Computational Workload Reduction}
\begin{figure}[h]
    \centering
    \includegraphics[scale=0.6]{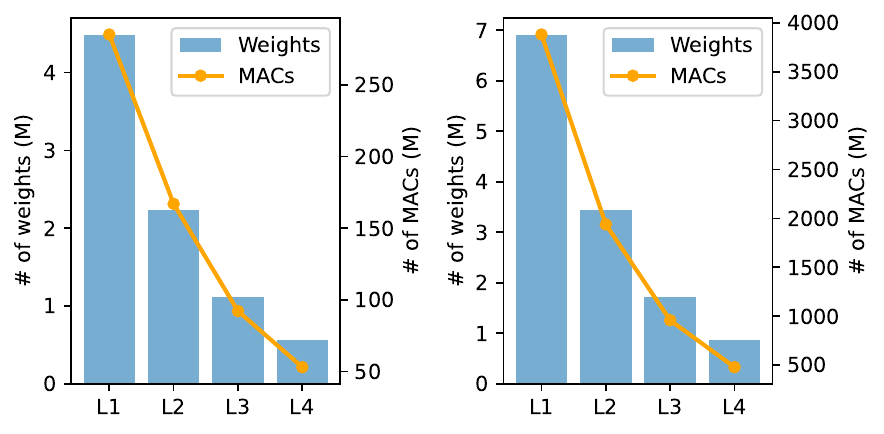}
    \caption{The effect of weight pruning on MACs reduction. \textbf{Left:} ResNet-18 on CIFAR-100. \textbf{Right:} WideResNet-22-8 on Tiny-ImageNet.}
    \label{fig:comp}
\end{figure}

Similar to previous work \cite{gong2022all}, we adopt Multiply-Accumulate operations (MACs) as a metric for assessing computational workload. Our strategy involves the implementation of weight pruning to effectively curtail MAC operations, thereby striving to conserve energy through reduced computational expenditures.
As shown in Fig.~\ref{fig:comp}, it becomes evident that there exists a strong correlation between the reduction in MACs and the density of weights. Specifically, the weight densities for layers L1 through L4 stand at [0.4, 0.2, 0.1, 0.05] respectively. This alignment underscores our ability to proficiently manage computational workload by means of adjusting weight density.

\section{Original ReLU Count Comparison}
\begin{figure*}[t]
    \centering
      \includegraphics[width=.95\linewidth]{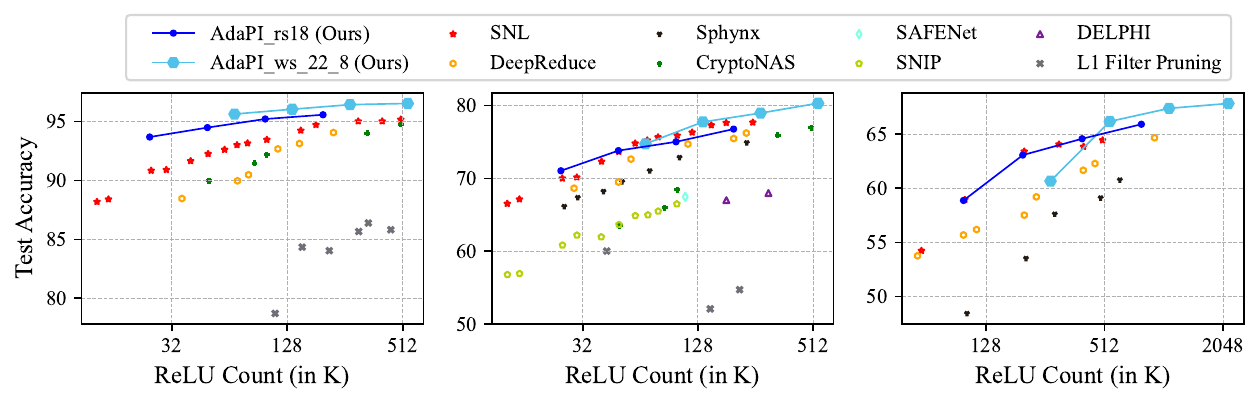}
  \captionof{figure}{\ouralg achieves Pareto frontiers of ReLU counts vs. test accuracy on \textbf{CIFAR-10 (left)}, \textbf{CIFAR-100 (middle)}, and \textbf{Tiny-ImageNet (right)} compared to the SOTA methods, where AdaPI\_rs18 and AdaPI\_ws\_22\_8 denote \ouralg on ResNet-18 and WideResNet-22-8, respectively. In each figure, the ReLU/weight density of \ouralg from left to right is L4 (0.05), L3 (0.1), L2 (0.2), and L1 (0.4).
  }
    \label{fig:Pareto_frontiers}
\end{figure*}

Given that prior research primarily concentrates on optimizing the balance between original (unnormalized) ReLU counts and test accuracy, we extend our evaluation to include this aspect in our work. As depicted in Fig.~\ref{fig:Pareto_frontiers}, our approach achieves Pareto frontiers concerning ReLU counts versus test accuracy when compared to other SOTA methods.
Furthermore, on CIFAR-100, our method achieves comparable test accuracy to SNL with a single set of weights, even though SNL optimizes performance for distinct ReLU budgets. Additionally, we achieve lower weight density compared to SNL by choosing different weight masks, corresponding to 0.05, 0.1, 0.2, and 0.4, respectively.

\begin{figure}[h]
    \centering    \includegraphics[width=.75\linewidth]{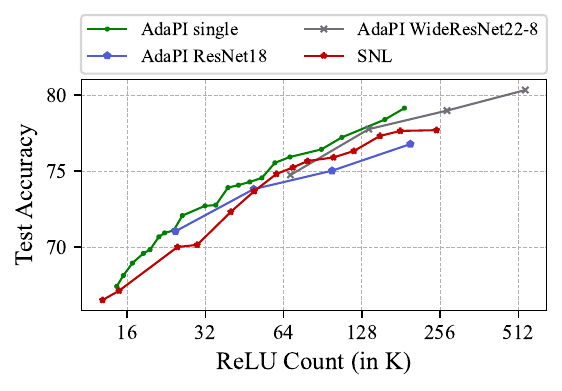}
    \caption{Performance comparison (ReLU count vs. text accuracy) of \ouralg-single on CIFAR-100.}
    \label{fig:wo_adp}
\end{figure}

We also provide a detailed comparison between \ouralg-single, \ouralg, and SNL in terms of ReLU budgets, as depicted in Fig.~\ref{fig:wo_adp}. Through model optimization for each mask pair ($M^b_r$ and $M^b_w$), \ouralg-single slightly outperforms \ouralg, albeit at the expense of storage efficiency.

It's worth noting that Fig.~\ref{fig:Pareto_frontiers} and Fig.~\ref{fig:wo_adp} specifically illustrate ReLU reduction, although test accuracy is also influenced by MAC reduction, which contributes to the adaptivity to varied energy budgets as well.

\bibliographystyle{named}
\bibliography{ijcai24}